\documentclass[runningheads]{llncs}
\usepackage{graphicx}
\usepackage{amsmath}
\usepackage{listings}
\usepackage{xcolor}
\usepackage{diagbox}

\lstdefinestyle{jsonStyle}{
  basicstyle=\ttfamily\small,
  breaklines=true,
  frame=single,
  columns=fullflexible,
  showstringspaces=false,
  commentstyle=\color{gray}\upshape,
  morestring=[s]{"}{"},
  morecomment=[l]{//}
}
\begin{document}
\title{Self-Agreement: A Framework for Fine-tuning Language Models to Find Agreement among Diverse Opinions}
\author{Shiyao Ding\inst{1}\and
Takayuki Ito\inst{1}}

\institute{Kyoto University, Kyoto-shi, Kyoto 606-8501, Japan \\ \email{ding@i.kyoto-u.ac.jp, ito@i.kyoto-u.ac.jp} \\}
\maketitle              % typeset the header of the contribution
\begin{abstract}
Finding an agreement among diverse opinions is a challenging topic in multiagent systems. Recently, large language models (LLMs) have shown great potential in addressing this challenge due to their remarkable capabilities in comprehending human opinions and generating human-like text. However, they typically rely on extensive human-annotated data.
In this paper, we propose Self-Agreement, a novel framework for fine-tuning LLMs to autonomously find agreement using data generated by LLM itself. Specifically, our approach employs the generative pre-trained transformer-3 (GPT-3) to generate multiple opinions for each question in a question dataset and create several agreement candidates among these opinions. 
Then, a bidirectional encoder representations from transformers (BERT)-based model evaluates the agreement score of each agreement candidate and selects the one with the highest agreement score. 
This process yields a dataset of question-opinion-agreements, which we use to fine-tune a pre-trained LLM for discovering agreements among diverse opinions.
Remarkably, a pre-trained LLM fine-tuned by our Self-Agreement framework achieves comparable performance to GPT-3 with only 1/25 of its parameters, showcasing its ability to identify agreement among various opinions without the need for human-annotated data.
\end{abstract}
\section{Introduction}
Consensus refers to a significant portion of a group agreeing on a specific topic, which is essential for collaboration and a crucial element of democracy. 
Consequently, consensus building as an important topic of multiagent systems has been extensively studied, such as finding an agreement through discussions among diverse opinions \cite{gu2021case}, which is often challenging \cite{bakker2022fine}.
The recent rise and success of large language models (LLMs), such as the generative pre-trained transformer-3 (GPT-3) \cite{brown2020language}, offer promising opportunities for addressing this challenge, by leveraging their capabilities in comprehending human opinions and generating human-like text.

Currently, employing LLMs for finding agreement among diverse opinions can be divided into two kinds of approaches. The first approach involves zero-shot learning on a pretrained LLM, such as GPT-3/4, which can output agreement without additional training. However, accessing these models requires the use of an application programming interface (API), which can introduce latency, high costs, and limited accessibility for research. 
Moreover, while some pretrained LLMs like Meta's LLaMA \cite{touvron2023llama} are available for local download, their massive parameters often lead to substantial storage and computational requirements during inference, making them less feasible for certain applications.

The second approach focuses on few-shot learning, which entails fine-tuning a LLM using a dataset containing opinions and agreement data. One notable example is \cite{bakker2022fine}, which relies on a 70 billion parameter model and expert-annotated data. 
However, this dependence on high-quality, human-generated data and the associated costs can be a significant barrier for many researchers and organizations seeking to leverage these models for consensus-building tasks.
These limitations highlight the need for more accessible and cost-effective solutions in applying LLMs to find agreement among diverse opinions without heavily relying on human resources.

In this paper, we address the above challenges by proposing Self-Agreement, a framework designed to autonomously find agreement among diverse opinions using LLMs. Our method consists of the following four key steps.
 First, we employ GPT-3 to generate multiple opinions on each question given a dataset containing various questions.
Next, we use GPT-3 to generate several potential agreements among these diverse opinions for each question.
We then evaluate the agreement score of each agreement candidate, choosing the one with the highest score as the optimal agreement to construct a question-opinion-agreement dataset.
Finally, we fine-tune a pre-trained LLM using the above dataset. This step allows the model to adapt to the specific task of finding agreements among various opinions.
 In our evaluation, we fine-tuned a 7 billion pre-trained LLM using our Self-Agreement framework. Remarkably, our model achieves a similar performance to GPT-3 with only 1/25 of its parameters, showcasing its ability to identify areas of agreement among conflicting views.

In summary, our contributions include:
1)  a large dataset consisting of various questions, opinions, and agreement candidates, which can serve as a valuable resource for building and evaluating consensus-building models;
2) the Self-Agreement framework which is designed to autonomously find agreement among diverse opinions using LLMs without human-annotated data;
3) a demonstration of fine-tuning a pre-trained model, with a comprehensive set of experiments confirming its effectiveness and performance.

\section{Related Work}
The concept of agreements can vary depending on the task. In this paper, we consider agreement as a form of opinion summarization from users, which may be agreed upon by all the users, similar to the approaches used in \cite{bakker2022fine} \cite{suhara2020opiniondigest}.

\textbf{Opinion Summarization}
Most opinion summarization methods follow a three-step process \cite{hu2004mining}\cite{Kim2011Comprehensive}: First, \textit{aspect extraction} is performed, which involves identifying the relevant features or aspects of the product that the user is commenting on. Second, \textit{sentiment prediction} is used to determine the sentiment of the extracted aspects, whether it is positive, negative, or neutral. This step helps to understand the overall opinion of the user. Finally, \textit{summary generation} is used to present the identified opinions to the user in a concise and easily understandable manner. 
This step involves condensing the extracted aspects and their corresponding sentiments into a brief summary. Most methods rely on extractive techniques for creating textual summaries, which select representative segments from the source text. However, this can result in loss of information that may be useful depending on user needs.

Most current approaches for opinion summarization, as described in the reference \cite{rush2015neural}, involve encoding documents and then decoding the learned representations into an abstractive summary. These methods leverage the success of sequence-to-sequence neural network architectures and are trained using sets of opinions and their corresponding summaries. In this approach, there is no need to explicitly identify aspects and sentiment for the opinion summarization task, as these are learned implicitly from the training data.
However, due to memory limitations, training these models end-to-end with a large number of input reviews for each target entity is practically infeasible \cite{Amplayo2021Informative}.

\textbf{LLMs for Opinion Summarization}
With the rapid development of LLMs, especially transformer-based LLMs, they have shown great potential for tackling tasks such as opinion summarization. The transformer model, proposed in 2017 \cite{vaswani2017attention}, has become a classic model for natural language processing (NLP) tasks, and various Transformer-based large language models have been proposed.

Bidirectional encoder representations from transformers (BERT) was one of the first models that pretrained on large corpora and fine-tuned for specific tasks, enabling it to capture intricate language patterns \cite{devlin2018bert}. 
Then, GPT have also seen significant advancements, with GPT-1, GPT-2 \cite{radford2019language}, and GPT-3 \cite{brown2020language}, focusing on large-scale unsupervised learning to generate human-like text. Other notable LLM architectures include RoBERTa \cite{liu2019roberta}, which improves upon BERT with enhanced pretraining and training techniques; Text-to-Text Transfer Transformer (T5) \cite{raffel2020exploring}, which adopts a unified text-to-text approach for various NLP tasks; and Large Language Model Meta AI  (LLaMA) \cite{touvron2023llama}.

Those LLMs have shown great potential in opinion summarization tasks, largely due to their ability to process and generate natural language. Bakker et al. \cite{bakker2022fine} fine-tuned a 70 billion parameter pretrained LLM to produce statements that maximize the expected approval for groups with diverse opinions. 
The model demonstrated exceptional performance, generating consensus statements preferred by over 70\% of human users compared to prompted LLMs. However, their approach relies on human annotations.
Although the Self-Instruction framework proposed by Wang et al. \cite{wang2022self} utilizes pre-trained LLM outputs for fine-tuning LLMs to follow human instructions, it still requires the preparation of seed instructions. 
Our Self-Agreement framework, on the other hand, specifically targets consensus-building tasks and does not depend on any human-generated instructions. This distinction highlights the advantages of the Self-Agreement framework in handling diverse opinions and generating agreement statements without extensive reliance on human-annotated data.

\section{Method}
Annotating and generating large-scale agreement data usually require significant human resources, as it often involves thousands of human-written questions and agreement data. Although those data  can be automatically output  in some cases, human input is still needed to score or rank the outputs, allowing LLMs to fit human preferences.
In this section, we present our Self-Agreement framework, where the entire process, from generating to scoring agreements, does not require human intervention. This makes it a completely automatic process, as shown in Fig.1.

\begin{figure}
\centering
\includegraphics[width=12cm]{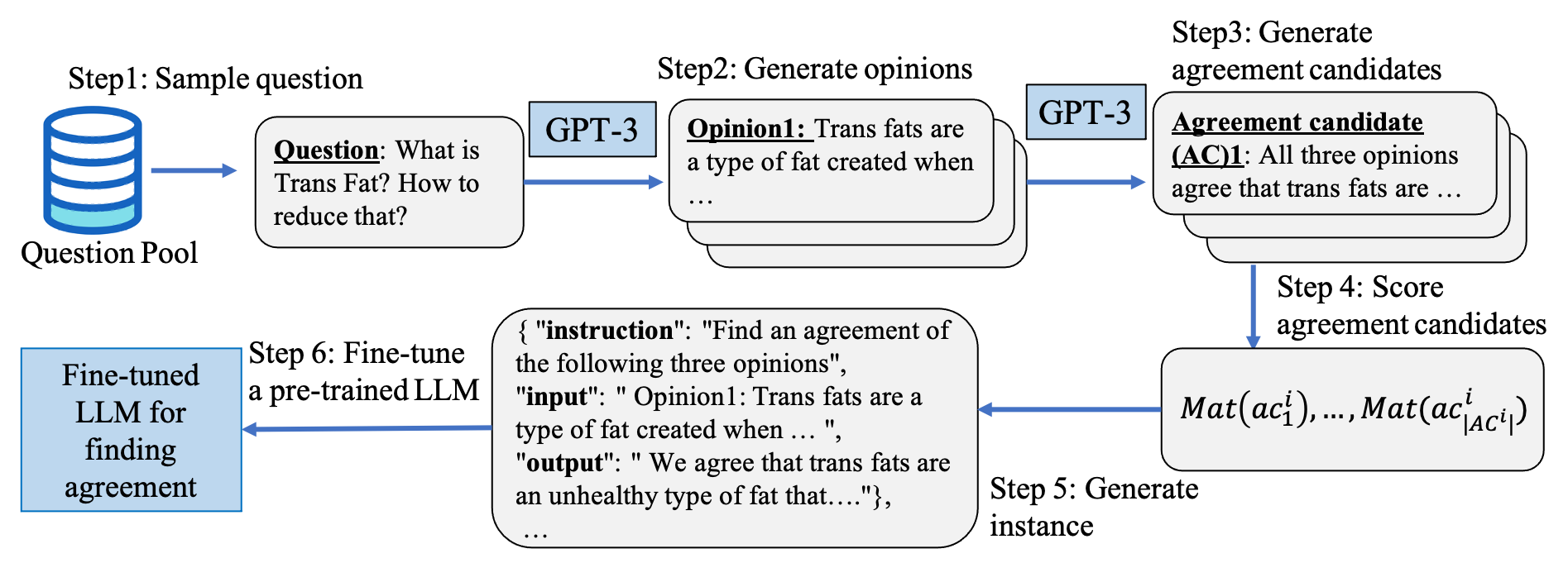}
\caption{The framework of the Self-Agreement.} \label{fig1}
\end{figure}

\subsection{Automatic Opinion-Agreement Data Generation}
We first define a consensus-building  instance $CB^i$ for question $i$ as a tuple of $CB^i=<q^i, OP^i, AC^i>$, where $q^i$ denotes question $i$, $OP^i=\{ op^i_1, ..., op^i_{|O^i|} \}$ is the opinion set, and $AC^i=\{ ac^i_1, ..., ac^i_{|AC^i|} \}$ is the set of agreement candidates. 
As mentioned in the previous section, we consider the definition of agreement as a form of opinion summarization from users, which may be agreed upon by all users. Then, for a set $Q=\{ q^1, q^2, ... , q^{|Q|} \}$ of questions, we aim to generate $|Q|$ consensus-building instances as a question-opinion-agreement dataset.

First, for opinion generation for each question $q^i$, inspired by the Self-Instruct approach, which utilizes large pretrained language models to create new and diverse instructions in a bootstrapping fashion, our framework employs the GPT-3 to generate several diverse opinions with the prompt ``Generate $n$ opinions for the question of $q^i$'', as shown in Steps 1 and 2 of Fig. 1. However, unlike Self-Instruction, which first relies on a  set of seed human-written opinions, Self-Agreement does not require any previously written human opinions.
Then, in Step 3 of agreement generation, we also employ GPT-3 to create $m$ agreement candidates with the prompt ``Find an agreement for the following opinions $OP^i$,'' where an instance is shown in Table 1.

\subsection{Score Agreement Candidates }
For each consensus-building instance, the goal is to choose the optimal agreement candidate that matches all the opinions well. 
First, we can define a scoring function $Mat: OP^i \times AC^i \rightarrow \mathcal{R}$ to evaluate how well agreement $ac^i_j$ matches opinion $op^i_k$.
In this paper, we set the range of values of $Mat$ as $[0, 1]$, where $Mat(op^i_k, ac^i_j)=0$ represents the agreement $ac^i_j$ being completely irrelevant to opinion $op^i_k$, and 1 means the agreement $ac^i_j$ completely supports opinion $op^i_k$. such as, $Mat(op^i_k, ac^i_j)|_{ac^i_j=op^i_k}=1$.

For each question $q^i$, we obtain $|OP^i|$ opinions and $|AC^i|$ agreement candidates. The model assigns a score to each candidate agreement, reflecting its relevance and compatibility with the given opinions. 
By calculating the agreement candidates according to their cumulative scores, we can identify the optimal agreement candidate that represents the various viewpoints. The best agreement candidate, $ac^i_{*}$, is the one with the highest agreement score:
\begin{equation}
    ac^i_{*}=\text{argmax}_{ac^i_j \in AC^i} \sum_{op^i_k\in OP^i}Mat(op^i_k, ac^i_j )
\end{equation}
Although various functions can be used as the agreement score function, in this paper, we employ a BERT-based model to evaluate the similarity between each opinion and agreement pair. This model computes contextualized embeddings for each sentence, capturing their semantic representations. The similarity score is then calculated by comparing these embeddings, typically using a distance metric such as cosine similarity. Thus, we take $ac^i_{*}$ as the output of the instance, as shown in Step 5 of Fig.1.

\begin{table}[h]
\centering
\caption{A sample of self-agreement process with three opinions and two candidate agreements on the topic of trans fats.}
\begin{tabular}{ p{2cm}  p{10cm} }
\hline
\textbf{Topic} & What is Trans Fat? How to reduce that? \\
\hline
\hline
\textbf{Opinion 1}& Trans fats are a type of fat created when hydrogen is added to vegetable oil during the production process, and can increase the risk of heart disease and other health issues. The best way to reduce trans fat intake is to eat a balanced diet with plenty of fruits, vegetables, whole grains, and lean proteins.   \\
\hline
\textbf{Opinion 2}& Trans fats are unhealthy and should be avoided whenever possible. To reduce trans fat intake, individuals should read labels carefully and opt for foods with no trans fat or low amounts of trans fat. Additionally, it is important to limit processed foods and choose items that are made with healthier fats, such as olive oil or avocado oil.  \\
\hline
\textbf{Opinion 3}& Trans fats are an unhealthy type of fat and should be avoided. To reduce trans fat intake, individuals should limit their consumption of processed and fast foods and instead focus on eating fresh, whole foods. Additionally, they should check food labels to make sure they are not consuming hidden trans fats.  \\
\hline
\hline
\textbf{Agreement Candidate 1}& All three opinions agree that trans fats are unhealthy and should be avoided. To reduce trans fat intake, individuals should limit their consumption of processed and fast foods, read labels carefully, and opt for foods with no trans fat or low amounts of trans fat. Additionally, it is recommended to focus on eating fresh, whole foods, and choosing items that are made with healthier fats, such as olive oil or avocado oil.\\
\hline
\textbf{Agreement Candidate 2}& All three opinions agree that trans fats are an unhealthy type of fat that should be avoided. To reduce trans fat intake, individuals should limit their consumption of processed and fast foods, check food labels, and opt for foods with no trans fat or low amounts of trans fat. Additionally, they should focus on eating fresh, whole foods with healthier fats, such as olive oil or avocado oil.\\
\hline
\end{tabular}
\end{table}

\subsection{Fine-tuning the Language Model}
After building the question-opinion-agreement dataset, we fine-tune a pre-trained language model for the task of consensus building, as shown in Step 6 of Fig.1. 
In this paper, we employ a 7-billion-parameter LLaMA model based on the Alpaca-LoRA architecture for fine-tuning. The Alpaca-LoRA architecture comprises two components: 1) the original 7-billion-parameter model, and 2) an adapter module. During the fine-tuning process, only the adapter module's parameters are updated, while the original model's parameters remain unchanged. Both components contribute to the inference process.

Each instance in the training dataset consists of an ``Instruction," an ``Input" and an ``Output", as shown in Fig.1. Focusing on the task of consensus building, we set the instruction as ``Find an agreement among the following opinions." This fine-tuning process equips the language model with the capability to efficiently identify consensus among diverse opinions.

\section{Evaluation}
\subsection{Evaluation Setting}
\textbf{Dataset} In this paper, we use Yahoo! Answers topic classification dataset \cite{zhang2015character} which includes 1,400,000 training samples and 60,000 testing samples. Each sample includes  following 5 parts: id, topic (class label) question title, question content and best answer, where the table A.1 in Appendix shows some examples.
We choose 1000 question titles from training samples for generating training dataset.
We then use each question content to generate opinions by using GPT-3. We consider both of the opinions have conflict and not, where the corresponding prompts are as follows,
prompt1:\textit{ Generate three opinions for the topic of $topic_i$ which do not have a conflict} and prompt2:
\textit{Generate three opinions for the topic }of $topic_i$ which  have a conflict. 
Then we use GPT-3 to generate an agreement by inputing the prompt as \textit{Please generate an agreement of the following opinions.} independently for $m$ times. Thus, for each question, we have $n$ opinions and $m$ agreement candidates. We set $q=1000$, $m=3$ and $n=4$ in this paper, which corresponds to 1000 questions, 6000 opinions (with conflict:3000, without conflict:3000) and 8000 agreement candidates.\\

\noindent\textbf{Fine-tuning LLM} For fine-tuning the LLM, we select a 7 billion model from LLaMA \cite{touvron2023llama} as a pre-trained model to be fine-tuned. We utilized the above dataset containing opinions with and without conflicts. Additionally, we divided it into two datasets based on the method of choosing an agreement candidate. We considered two approaches: selecting an optimal agreement candidate or randomly picking one for training. Consequently, this generated four distinct cases, as shown in Table 2.

\begin{table}[htbp]
\centering
\begin{tabular}{|p{2cm} | p{5cm} |p{5cm} |}
\hline
 & Random Agreement Candidate & 
 Optimal Agreement Candidate \\
\hline
Without conflict opinions & 
\begin{minipage}{5cm}
\textbf{Prompt}: Generate three opinions for the topic which do not have a conflict \\
\textbf{Output}: Randomly choosing one from all agreement candidates as Output 
\end{minipage}
& \begin{minipage}{5cm}
\textbf{Prompt}: Generate three opinions for the topic which do not have a conflict \\
\textbf{Output}: Optimally choosing one from all agreement candidates as Output 
\end{minipage}  \\
\hline
With conflict opinions & 
\begin{minipage}{5cm}
\textbf{Prompt}: Generate three opinions for topic which have a conflict \\
\textbf{Output}:Randomly choosing one from all agreement candidates as Output. 
\end{minipage}
& \begin{minipage}{5cm}
\textbf{Prompt}: Generate three opinions for topic which have a conflict \\
\textbf{Output}: Optimally choosing the one with maximized consensus score from all agreement candidates as Output 
\end{minipage} \\
\hline
\end{tabular}
\caption{Four cases to fine-tune LLMs.}
\end{table}

Correspondingly, we employ GPT-3 as a baseline for comparison. For the test set, we randomly select 100 questions from the same dataset, excluding those used in the training dataset, to generate both opinions with conflicts and without conflicts. Table A.2 provides some examples of these test cases.

\begin{figure}
\centering
\includegraphics[width=12cm]{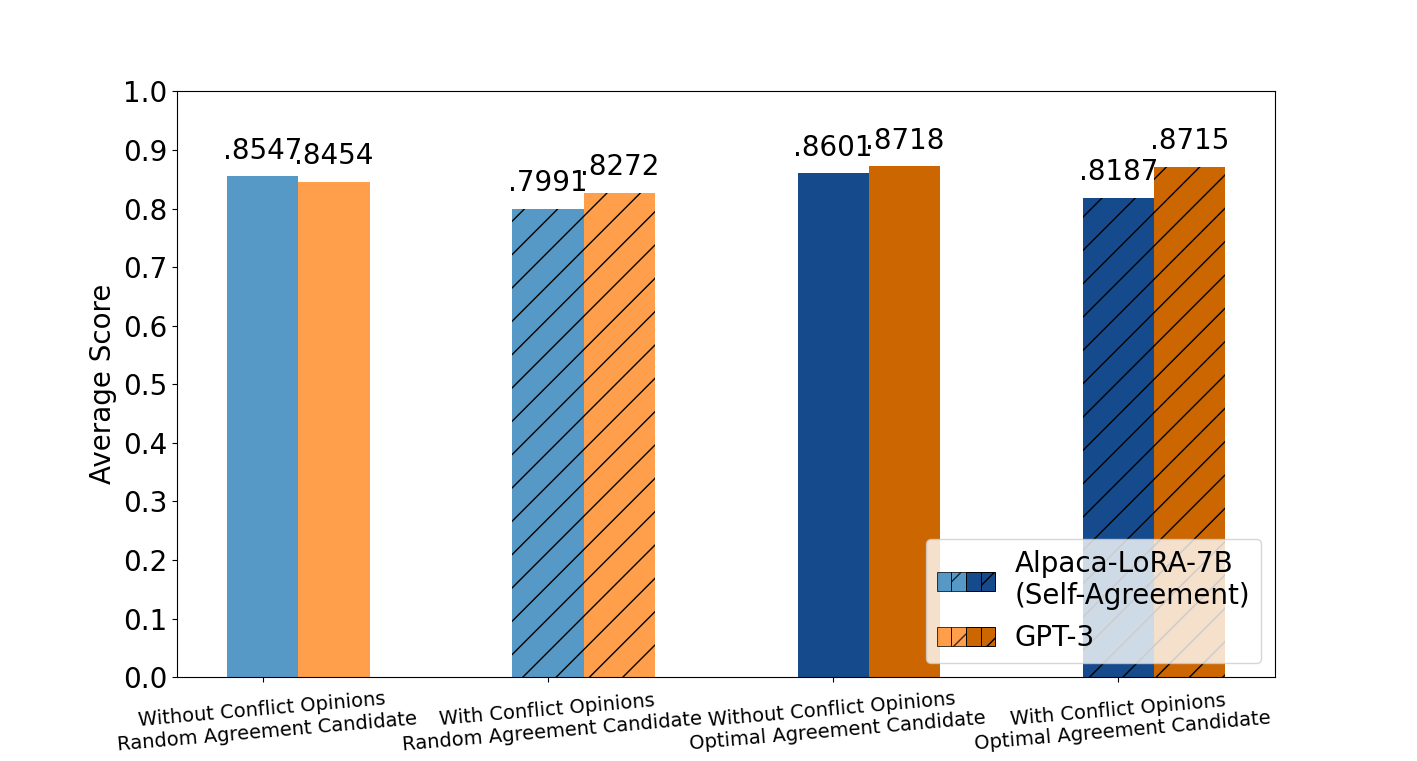}
\caption{The comparison of finding agreement results between GPT-3 and Alpaca-LoRA-7B (Self-Instruct) in four cases. To calculate the average agreement score, agreements are generated from the 100 test samples, and each agreement score is calculated using the summation component in Eq. (1). } \label{fig1}
\end{figure}

\subsection{Evaluation Results}
We compare our methods with GPT-3 and present the results for the four cases depicted in Fig. 2.
To calculate the average agreement score, agreements are generated from the 100 test samples, and each agreement score is calculated using the summation component in Eq. (1).
First, when randomly selecting an agreement candidate as the final opinion summarization, Alpaca-LoRA-7B (Self-Instruct) achieves comparable results in both cases of opinions with conflicts and without conflicts.
Also, both Alpaca-LoRA-7B (Self-Instruct) and GPT-3 exhibit lower scores for opinions with conflict than for opinions without conflict.
This can be attributed to the fact that it is generally more challenging for agreement candidates to accommodate conflicting opinions.

In the case where we select one optimal agreement candidate, the scores for both models improve. For instance, in the case of Alpaca-LoRA-7B (Self-Instruct), the score increases to 0.86 from 0.85 in the situation without conflicting opinions and to 0.81 from 0.79 in the situation with conflicting opinions.
These improvements highlight the benefits of selecting the optimal agreement candidate for opinion summarization tasks.
Most notably, Alpaca-LoRA-7B (Self-Instruct) demonstrates competitive performance in opinion summarization when compared to GPT-3, which has 175 billion parameters.
This underscores the effectiveness of our Self-Agreement framework to fine-tune an LM model in handling diverse opinions.

\begin{table}[h]
\centering
\caption{One instance of comparing Agreement candidates of GPT-3 and Alpaca-LoRA-7B (Self-Instruct) from opinions with conflict.}
\begin{tabular}{ p{2cm}  p{9cm}   }
\hline
\textbf{Topic} & Which is the country with top development indicators?\\
\hline
\hline
\textbf{Opinion 1}& India is the country with top development indicators because it has achieved great progress in the fields of education, health, and poverty alleviation.  \\
\hline
\textbf{Opinion 2}& China is the country with top development indicators because it has made great strides in economic growth and infrastructure development. \\
\hline
\textbf{Opinion 3}& The United States is the country with top development indicators because it has the highest GDP per capita and the most advanced technology.\\
\hline
\hline
\textbf{Agreement  Candidate (GPT-3)}&  We agree that India, China, and the United States are all countries with top development indicators. India has achieved great progress in the fields of education, health, and poverty alleviation; China has made great strides in economic growth and infrastructure development; and the United States has the highest GDP per capita and the most advanced technology. Together, these three countries demonstrate the highest level of development in the world.\\
\hline
\textbf{Agreement Candidate \quad (Self-Agreement)}&  We agree that India, China, and the United States are all countries with top development indicators. India has achieved great progress in the fields of education, health, and poverty alleviation, China has made great strides in economic growth and infrastructure development, and the United States has the highest GDP per capita and the most advanced technology. \textit{\textbf{However, we disagree on which country has the top development indicators, as each country has its own strengths and weaknesses.}} \\
\hline
\end{tabular}
\end{table}

Further, we list an instance of the above result.
Table 3 shows the comparison results of the agreement candidates generated by GPT-3 models and Alpaca-LoRA (Self-Agreement), in response to a given topic: \textit{Which is the country with top development indicators?}.
The opinions represent three different views on which country has the top development indicators.
In the GPT-3 generated agreement candidate, the model introduces India, China, and The United States as countries with high development indicators. Although these countries are well-regarded for their development, the response deviates from the given opinions.
This indicates that GPT-3 may not effectively capture the essence of the conflicting opinions in this scenario.
However, the Alpaca-LoRA-7B (Self-Instruct) generated agreement candidate acknowledges all three opinions and successfully incorporates the key points from each. It emphasizes the strengths of India, China, and the United States while also recognizing that it is difficult to determine which country has the top development indicators, as each has its own strengths and weaknesses (shown in the highlighted parts of Table 3). This response demonstrates the effectiveness of the Self-Agreement framework in accurately capturing and representing diverse opinions in a conflict situation.

Both GPT-3 and the Self-Agreement framework demonstrate their ability to synthesize diverse opinions into a coherent agreement candidate. However, the Alpaca-LoRA-7B model fine-tuned by Self-Agreement framework provides a more balanced perspective by acknowledging the underlying disagreement and avoiding a definitive conclusion. This illustrates the effectiveness of the Self-Agreement framework in handling conflicting opinions and generating a consensus statement that fairly represents the diversity of viewpoints.

Further, catastrophic forgetting is a phenomenon in which a model trained on a new task, significantly degrades its generalization performance on the original task, resulting in a severe loss of previous knowledge. To test for catastrophic forgetting, we use some instruction data from the Self-Instruction framework and find that the Alpaca-LoRA-7B model, after fine-tuning with the Self-Agreement framework, still demonstrates a good response to general instructions. Some instances are listed in Table A.2.

\section{Conclusion}
In this paper, we introduced the Self-Agreement, an efficient framework to fine-tune LLMs to autonomously find agreement among diverse opinions. Our method eliminates the need for expensive human-generated data. 
We also presented a large dataset of questions, opinions, and agreement candidates, serving as a valuable resource for future consensus-building models. Our experiments highlight the effectiveness of our framework in consensus-building tasks while achieving comparable performance to GPT-3 with only 1/25 of its parameters.

\bibliographystyle{splncs04}
\bibliography{self-agreement.bib}

\section*{Appendix}
The Fig.A.1 illustrates the topic distribution of questions from both train and test samples. As can be observed from the pie charts, the questions are distributed across various topics. Moreover, Table A.1 provides several examples from the Yahoo dataset, showcasing the diversity of questions and answers across different topics.
\renewcommand{\thefigure}{A.\arabic{figure}}
\setcounter{figure}{0}

\begin{figure}
\centering
\includegraphics[width=12.5cm]{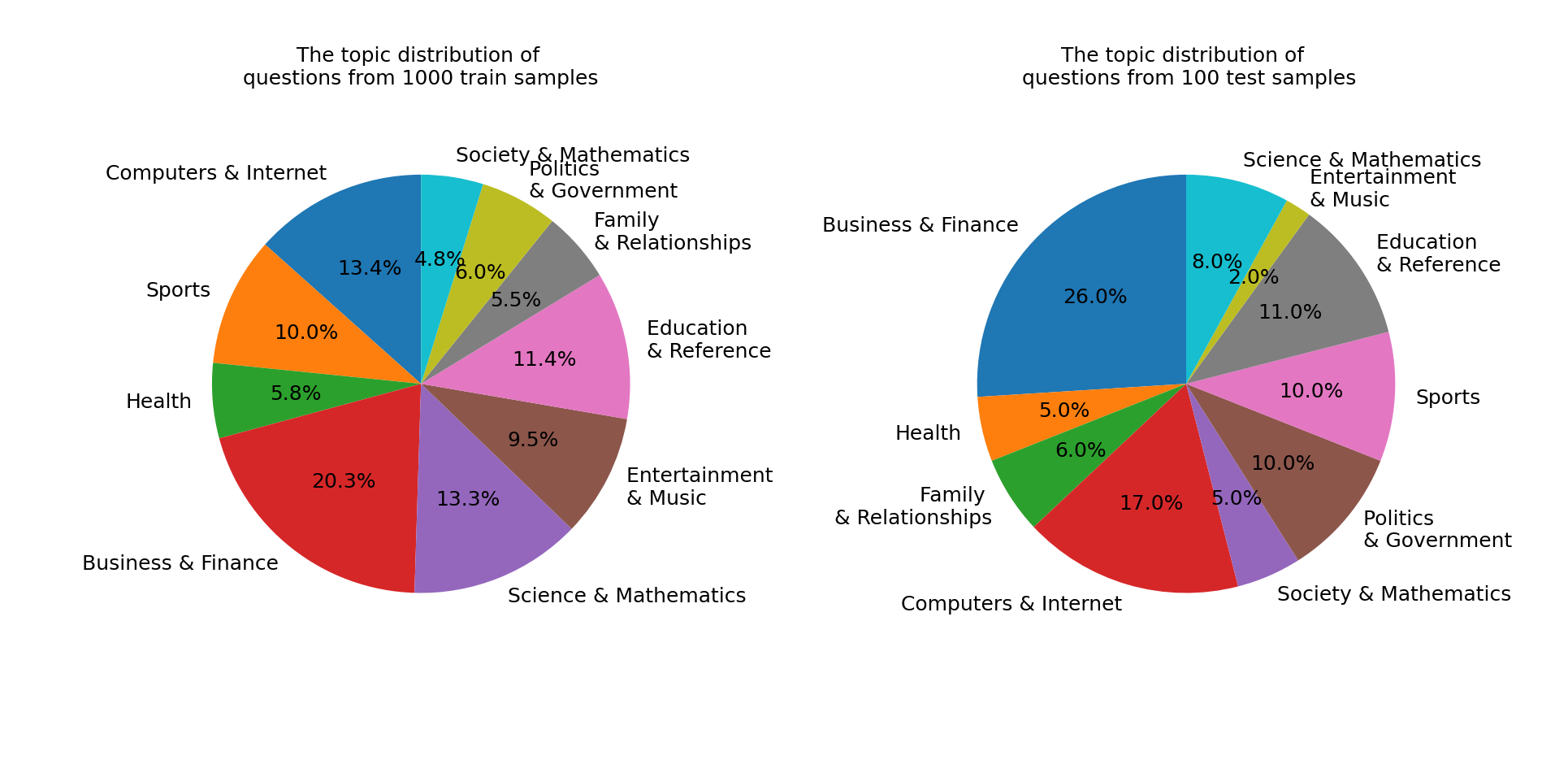}
\caption{The topic distribution of questions from train and test samples.} \label{fig1}
\end{figure}

\renewcommand{\thetable}{A.\arabic{table}} 
\setcounter{table}{0}
\begin{table}[htbp]
\centering
\begin{tabular}{|p{0.5cm} | p{1.2cm} |p{3cm} |p{3.5cm} |p{3.5cm} |}
\hline
\textbf{Id} & \textbf{Topic (Class label) }& 
\textbf{Question Title }& \textbf{Question Content }& \textbf{Best Answer} \\
\hline
0 & 4 (Computers  Internet)
& "why doesn't an optical mouse work on a glass table?" & "or even on some surfaces?" &
"Optical mice use an LED and a camera to rapidly capture images of the surface beneath the mouse. ... 
\\
\hline
1 & 5 (Sports) &  "What is the best off-road motorcycle trail ?" & "long-distance trail throughout CA" & "i hear that the mojave road is amazing!"\\
\hline
2 & 2 (Health) & "What is Trans Fat? How to reduce that?" & "I heard that tras fat is bad for the body. Why is that? Where can we find it in our daily food?" & "Trans fats occur in manufactured foods...\\
\hline
\end{tabular}
\caption{Some examples of Yahoo answer topic dataset}
\end{table}

The following table demonstrates how well the model of Alpaca-LoRA-7B can still perform general instructions after fine-tuning with the Self-Agreement framework, indicating that catastrophic forgetting has been mitigated.
\begin{table}[h]
\centering
\caption{Sample general instruction responses from Alpaca-LoRA-7B after fine-tuning with the Self-Agreement framework.}
\begin{tabular}{ |p{3cm}| p{3cm} | p{6cm} |  }
\hline
\textbf{Instruction} &\textbf{Input}  & \textbf{Output} \\
\hline
Give three tips for staying healthy.&  & 1. Eat a balanced diet of fruits, vegetables, and whole grains.
2. Exercise regularly to keep your body in shape.
3. Get enough sleep to recharge your body and mind.  \\
\hline
What are the three primary colors?& &The three primary colors are red, blue, and yellow. These colors can be used to create any other color by mixing them together. \\
\hline
Describe the structure of an atom.& &An atom is made up of a nucleus surrounded by a cloud of electrons. The nucleus is made up of protons and neutrons, while the cloud of electrons is made up of negatively charged electrons.\\
\hline
Classify the following into animals, plants, and minerals &   Oak tree, copper ore, elephant.&1.Oak tree is an example of an animal, as it is a living organism.
2.Copper ore is an example of a mineral, as it is a naturally occurring substance.
3.Elephant is an example of a plant, as it is a living organism that grows and reproduces.\\
\hline
Identify the odd one out.&Twitter, Instagram, Telegram &Telegram is the odd one out as it is a messaging app, while Twitter and Instagram are social media platforms.\\
\hline
\end{tabular}
\end{table}

\end{document}